\documentclass[10pt,twocolumn,letterpaper]{article}

\usepackage{cvpr}              %
\usepackage{tabularx}

\usepackage[dvipsnames]{xcolor}

\newcommand{\TODO}[1]{\textbf{\color{red}[TODO: #1]}}
\renewcommand{\TODO}[1]{}

\usepackage{amsmath}
\usepackage{amssymb}
\usepackage{mathtools}
\usepackage{amsthm}

\theoremstyle{plain}

\theoremstyle{definition}

\theoremstyle{remark}

\DeclareMathOperator*{\argmin}{arg\,min}

\usepackage{soul}
\usepackage{multirow}
\usepackage{caption}
\usepackage{subcaption}

\definecolor{cvprblue}{rgb}{0.21,0.49,0.74}
\usepackage[pagebackref,breaklinks,colorlinks,citecolor=cvprblue]{hyperref}

\def\paperID{*****} %
\def\confName{CVPR}
\def\confYear{2024}

\title{Non-Robust Features are Not Always Useful in One-Class Classification}

\author{Matthew Lau\\Georgia Tech \and Haoran Wang\\Georgia Tech \and Alec Helbling\\Georgia Tech \and 
Matthew Hull\\Georgia Tech \and ShengYun Peng\\Georgia Tech
\and Martin Andreoni\\Technology Innovation Institute
\and Willian T. Lunardi\\Technology Innovation Institute \and Wenke Lee\\Georgia Tech}

\begin{document}
\maketitle

\begin{abstract}
The robustness of machine learning models has been questioned by the existence of adversarial examples.
We examine the threat of adversarial examples in practical applications that require lightweight models for one-class classification.
Building on \citet{adv_ex_features_not_bugs}, we investigate the vulnerability of lightweight one-class classifiers to adversarial attacks and possible reasons for it.
Our results show that lightweight one-class classifiers learn features that are not robust (e.g. texture) under stronger attacks.
However, unlike in multi-class classification \citep{adv_ex_features_not_bugs}, these non-robust features are not always useful for the one-class task, suggesting that learning these unpredictive and non-robust features is
an unwanted consequence of training.
\end{abstract}    

\section{Introduction}

One-class classification, also known as anomaly detection, is useful in security applications.
For instance, drone detection is useful for law enforcement agencies to maintain the safety of airspaces.
However, devices that deploy automated detection may have computational and power constraints, limiting the application of computer vision models.
These practical concerns motivate the need to integrate lightweight models into one-class classifiers.

\begin{figure}
\vspace{-1mm}
    \centering
    \includegraphics[trim=0.5cm 16.6cm 0.5cm 4.875cm, clip, width=0.75\textwidth]{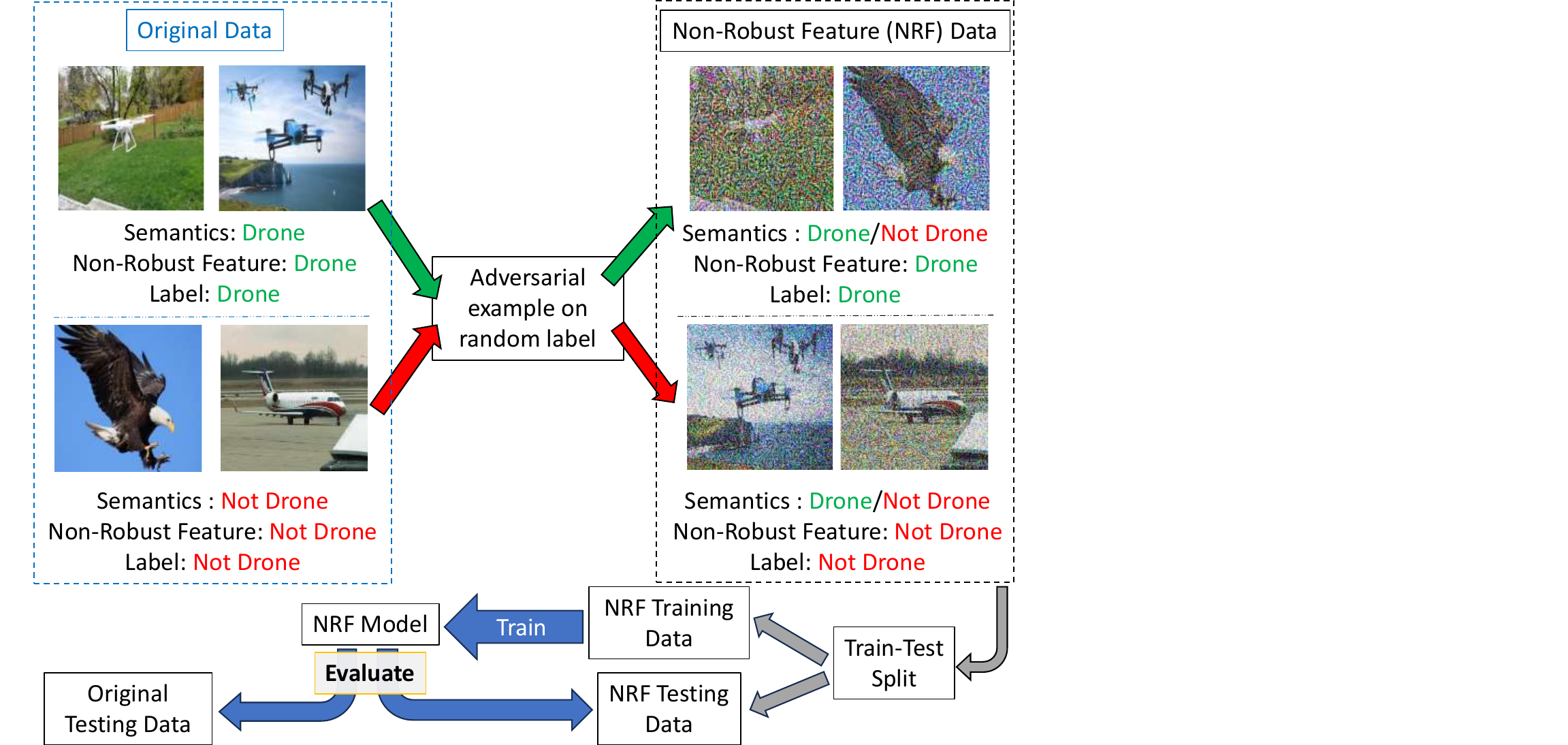}
    \caption{Evaluation framework of the usefulness of non-robust features (e.g. texture) on one-class classification, adapted from \citet{adv_ex_features_not_bugs}.}
    \label{fig:nrf_summary}
    \vspace{-2.5mm}
\end{figure}
However, computer vision neural network models can be vulnerable to adversarial attacks which try to evade or cause false detections.
To evaluate the security of deployed models, we investigate the usefulness and robustness of features learnt by lightweight one-class classifiers.
Our contributions are as follows:
\begin{enumerate}
    \item We test the \textbf{performance and vulnerability of lightweight one-class classifiers}. 
    We find that these models learn useful features, but their predictive power diminishes with smaller models and adversarial drift.
    Additionally, models are vulnerable to adversarial attacks regardless of adversarial drift, suggesting that they learn non-robust features (e.g., texture, background).
    \item We test a hypothesis from \citet{adv_ex_features_not_bugs} regarding the \textbf{reason} for adversarial examples, specifically that non-robust features are spuriously correlated with the semantics in the image used for the prediction task.
    Contrary to their findings in multi-class classification, we show that this is not the case in lightweight one-class classifiers -- such \textbf{non-robust features are not necessarily correlated with the semantics of the image useful for one-class classification}, and their correlation (or lack thereof) is independent of the size or adversarial robustness of the model.
\end{enumerate}
These findings suggest that, in addition to learning useful features, neural networks sometimes learn features that are not useful nor robust to adversarial attacks.
This calls for future work on avoiding learning such features.

\section{Background}

\subsection{Reasons for Adversarial Examples}

\citet{carlini_wagner_classifiers} demonstrate that multi-class classifiers are generally vulnerable to adversarial attacks, while \citet{carlini_wagner_detection} indicate that detection of these attacks can also be evaded.
Several works have been conducted to understand why neural networks are vulnerable.
A theory proposed by~\citet{adv_ex_features_not_bugs} suggests that neural networks learn non-robust features (NRFs) which are used for the task they are trained on.
For instance, the texture and background could be correlated with the semantics of the image, which introduces unwanted spurious correlations between the class label and the image texture and background~\cite{cv_bias}.
When the model learns these undesirable but predictive shortcuts, adversarial examples exist by intervening on these NRFs while keeping the semantics of the image the same.
\citet{adv_ex_not_real_features} builds on their results in multi-class classification to show that 
these NRFs are task-specific, transferring poorly to other tasks.

We aim to investigate the reasons for the existence of adversarial examples in one-class classification, which we view from an anomaly detection perspective.
One-class classification is different from~\citet{adv_ex_features_not_bugs} and~\citet{adv_ex_not_real_features} in two ways.

First, we consider distribution shift of anomalies, also known as \textit{adversarial drift} ~\citep{evasion_attacks_test_time, 1.5C_original_paper}, which is standard in anomaly detection~\citep{Unifying_Review_AD_Ruff}.
One-class classification relaxes the closed-world assumption that multi-class classifiers~\citep{adv_ex_features_not_bugs} usually have, which is that test-time semantics are present during training.
In one-class classification, the distribution of anomalies at test time is unknown and likely to differ from that at train time, potentially exhibiting multiple modes (e.g., many classes with different semantics).
Even as adversarial drift increases, meaning the distribution of test anomalies moves further away from the anomalies seen during training, we still aim for zero-shot generalization to classify these unseen classes as anomalies.
For instance, if birds and planes are anomalies during training, test-time anomalies might include jeeps and cranes. Instead of altering the loss function as suggested by \citet{adv_ex_not_real_features}, we modify the distribution of anomalies. For example, changing the loss function might involve altering the normal class within the same dataset, whereas changing the distribution 
alters the test-time anomalies from train-time anomalies.

Second, attacks in the one-class scenario are limited.
Ideologically, normal data can only be subjected to untargeted attacks, which aim to make the data appear less like the normal class to the classifier. 
Conversely, anomalous data can only be subjected to targeted attacks.
This one-class set-up is different from multi-class settings like \citet{adv_ex_features_not_bugs} where targeted attacks are launched.

\subsection{Set-up: Defining `Features'}
We first summarize the set-up of \citet{adv_ex_features_not_bugs}, which we adapt.
Define features $\mathcal F :=\{f:\mathcal X \to \mathbb R\}$ as the collection of all possible outputs of each neuron before the classifier layer $c$ (e.g., in the penultimate layer).
Features refer to abstract notions of attributes of an image used for prediction, such as the number of wings in an image.
A classifier $C := c\circ [f_i]_{i=1}^r$ has the classifier layer composed of $r$ features.
Its output is in $[-1,1]$. 
Denote the ground truth label for $\mathbf x$ as $y(\mathbf x)\in \{\pm1\}$, and suppress the dependence on $\mathbf x$ to have $y=y(\mathbf x)$.
Features are then split into 2 categories -- their utility and robustness with respect to a distribution $\mathcal D$:
\begin{itemize}
    \item \textbf{$\rho$-useful features $f$ on distribution $\mathcal D$} are features that are $\rho> 0$ correlated to the label on average:
    \begin{equation}
    \label{eqn:usefulness}
        \mathbb E_{\mathbf{x}\sim \mathcal D}[y\cdot f(\mathbf x)]\geq \rho.
    \end{equation}
    Correspondingly, features that are not useful have $\rho \approx 0$.
    \item \textbf{$\gamma$-robustly useful features $f$ on distribution $\mathcal D$} are features that are $\gamma> 0$ correlated to the label on average even under adversarial perturbations $\Delta$:
    \begin{equation}
        \mathbb E_{\mathbf{x}\sim \mathcal D}[\inf_{\delta \in \Delta} y\cdot f(\mathbf x + \delta)]\geq \gamma.
    \end{equation}
    \item \textbf{Useful non-robust features } are useful features where no $\gamma\geq 0$ exists for distribution $\mathcal D$ for threat model $\Delta$.
\end{itemize}

We generalize the NRF set-up in \citet{adv_ex_features_not_bugs} by looking at usefulness and robustness on different distributions, accounting for adversarial drift.

\subsection{Training on a non-robust feature dataset}
We adapt \citet{adv_ex_features_not_bugs} to generate a dataset with NRFs after training one-class classifier $C$ on the original dataset.
We aim to (1) destroy the correlation between the robust features (e.g., semantics) of the image and the label, and (2) increase the correlation between the NRFs and the label.
This way, the label in the NRF dataset represents the class corresponding to the respective NRFs, not the robust features.
\looseness=-1

To remove the correlation between robust features and the labels,
for each image (train and test), we randomly pick a label $y^{NRF}\in \{\pm1\}$ that will become its new label in the NRF dataset.
To increase the correlation between NRFs and the labels, we perform a targeted adversarial attack on the original classifier $C$ with label $y^{NRF}$ using perturbation
\begin{equation}
    \delta^* = \argmin_{\delta \in \Delta} \mathcal L(y^{NRF},C(\mathbf x+ \delta))
\end{equation}
where $\mathcal L$ is the loss function.
The attack ensures that each new image has NRFs correlated with the new label $y^{NRF}$.

Splitting this NRF dataset into training and testing, we train another model $C'$ on the training NRF dataset, which we refer to as an NRF model.
$C'$ can only learn correlations between NRFs.
If the NRFs are useful (i.e., correlated with the label), we expect good performance on both the NRF test dataset.
Otherwise, we expect random performance.

\section{Experiments}

\subsection{Experimental Set-Up}

We train a one-class classifier to distinguish drones (positives) from non-drones (negatives).
The classifier uses a pre-trained frozen encoder on ImageNet-1K and a trainable one-class classifier head.
In increasing size of the encoders, we test (1) MobileNetV3-small which 
has 3M parameters \citep{mobilenetv3_paper}, (2) EfficientNetB0 which 
has 5.3M parameters \citep{tan2020efficientnet}, and (3) ResNet18 which 
has 11M parameters \citep{resnet_paper,resnet18_param_count}.
To allow supervised training while maintaining the bias of enclosing decision boundaries for the desired (positive) class, we follow the approach outlined in \citet{lau2024OSR}.

We designate drones from the Drone vs. Bird dataset \citep{BirdVsDroneData} as the positive class. 
In practice, domain knowledge on anomalies of interest during deployment is often present, so relevant examples can be included during training.
Hence, we use flying objects for the negative class: birds from the Drone vs. Bird dataset, bald eagle and airliner images from ImageNet-1K \citep{imagenet15russakovsky}.
We introduce different distribution shifts beyond the known anomalies to simulate varying degrees of adversarial drift for unseen anomalies.
Tab.~\ref{tab:unseen_classes} categorizes these anomalies by two criteria: the imagery source and semantic similarity to the training data.
For the imagery source, in addition to testing on Drone vs. Bird and ImageNet-1K images, we also test on CIFAR-10 images \citep{cifar10}.
CIFAR-10 images are lower resolution than the training images, introducing one source of adversarial drift.
To introduce another source of adversarial drift, we add land vehicles for testing, such as jeeps, which are semantically different from flying objects.
We refer to individual classes like jeeps as negative subclasses of the overall negative class.
\looseness=-1

The performance of models on different negative subclasses corresponds to the usefulness of features for each subclass.
Our main metric will be the average precision (the area under the precision-recall curve) across 3 runs over each negative subclass and the overall negative class.
The average precision allows us to measure the separation between the positive drone class and any negative (sub)class in a more general setting without the specifics of setting a threshold.
We also observe similar patterns for other metrics (precision, recall, and F1 score under $90\%$ true positive rate) but leave them out due to space constraints.
 
\begin{table}[t]
    \footnotesize
\addtolength{\tabcolsep}{-2pt}
    \centering
    \caption{Adversarial drift of unseen anomalies, simulated with classes from ImageNet-1K and CIFAR-10.}
    \begin{tabularx}{\columnwidth}{X|X|X}
    \toprule
        Semantics $\backslash$ Imagery & Similar (ImageNet) & Different (CIFAR) \\
        \midrule
        Similar & Vultures, Airships & Airplanes, Birds \\
        \hline
        Different & Crane2, Jeeps & Trucks \\
        \bottomrule
    \end{tabularx}
    \label{tab:unseen_classes}
\end{table}

\subsection{Results}

\paragraph{Performance}
\begin{figure}
    \centering
    \includegraphics[width=\linewidth]{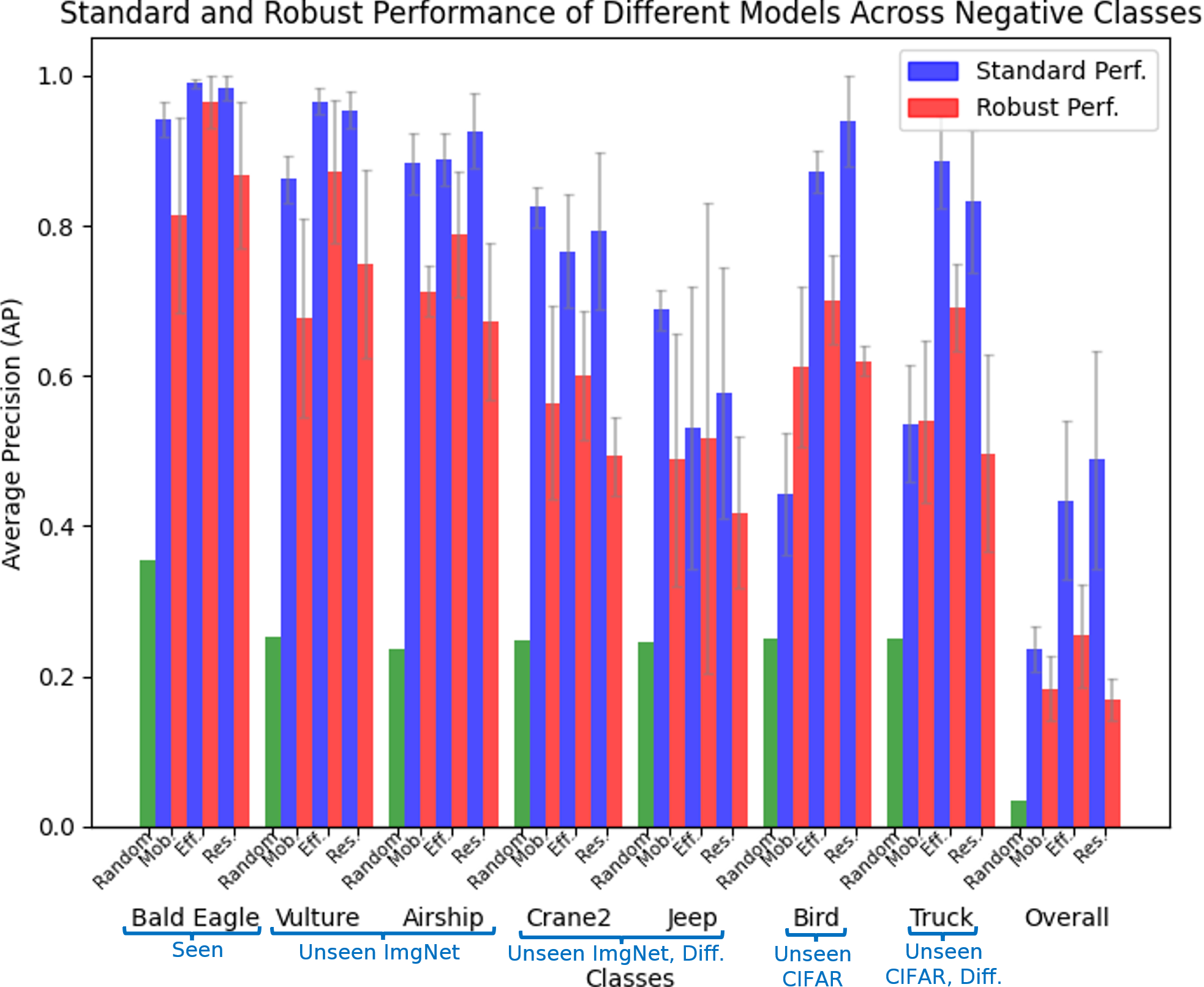}
    \caption{Standard and robust performance of different backbones across a subset of negative subclasses and the overall negative class.
    Each group of bars represents the average precision of the negative (sub)class against the positive drone class for different classifiers.
    The first green bar is the random baseline, followed by pairs of bars which are the standard and robust performance of MobileNetV3 (Mob.), EfficientNetB0 (Eff.) and ResNet18 (Res.) in order.
    Classes are grouped by roughly increasing adversarial drift from left to right, with semantic differences (diff.) and imagery differences (ImageNet-1K/CIFAR-10).}
    \label{fig:standard_perf_pgd_linf_025}
\end{figure}

The blue bars in Fig. \ref{fig:standard_perf_pgd_linf_025} shows our results on a subset of subclasses, and Tab. \ref{tab:drone_AUPR} in Appendix \ref{appendix:detailed results} has the full results.
For all (sub)classes, we see that all models achieve non-trivial performance, with all the blue bars being much higher than the green bars (which refer to random performance).
We observe that performance on seen classes is generally higher and has a lower variance for seen subclasses than unseen subclasses, suggesting that features learned for seen subclasses are very useful (i.e., $\rho$ is high, with little variance across runs).
A drop in performance for crane2 and jeep highlights that semantic adversarial drift affects performance the most, suggesting that the positive class (drone) semantics may not have been learned well.
In particular, MobileNetV3 (the smallest backbone model) has a drop in performance as adversarial drift increases on both semantics and imagery drifts, while EfficientNetB0 and ResNet18 backbones seem to produce more useful features for CIFAR-10 (imagery drift) but there is still some variance in their usefulness.
As expected, the usefulness of learned features tend to decrease with adversarial drift.

\paragraph{Vulnerability}
We evaluate the adversarial robustness of trained models by performing white-box adversarial attacks on these models.
This is known as robust performance, and it corresponds to the robustness of features learned.
We observe that stronger attacks are required, with the original $\ell_2$ projected gradient descent (PGD) attack with attack strength $\epsilon=0.25$ in \citep{adv_ex_features_not_bugs} not being effective and $\ell_\infty$ PGD with $\epsilon=4/255$ in \citep{adv_ex_not_real_features} being mildly effective.
We posit that the increased robustness of features learned could be due to the easier task of one-class classification compared to multi-class classification, but leave this to future work.
Hence, we report robust performance for $\ell_\infty$ PGD with $\epsilon=0.25$ as the red bars in Fig. \ref{fig:standard_perf_pgd_linf_025}.
With this stronger attack, every model can achieve almost random performance for every subclass, regardless of adversarial drift.
The ResNet18 backbone is the exception with consistently more robust performance than the other two, but it is also more computationally heavy.
Nevertheless, the effectiveness of the attack suggests that models learn NRFs during training.
We proceed to focus on attacks with strength $\epsilon\geq 0.25$ to explain adversarial vulnerability.

\paragraph{Usefulness of NRFs}
We create NRF datasets with $\ell_\infty$ PGD of strengths $\epsilon=0.5, 0.25$ and evaluate NRF models on NRF test data.
We also include attack strength $\epsilon=4/255$ to follow \citet{adv_ex_not_real_features}, but note that features are still relatively robust to this threat model.
Sample NRF images created with $\epsilon=0.5$ are shown in Fig. \ref{fig:nrf_summary}.
We report the overall average precision in Fig. \ref{fig:NRF_Model_on_NRF_data} and subclass results in Tab. \ref{tab:drone_AUPR} in Appendix \ref{appendix:detailed results}.
An interesting result emerges.
MobileNetV3 and ResNet18 models trained on the NRF training data have close to perfect performance on $\epsilon=0.5, 0.25$ NRF test data, suggesting that NRF features from these models are correlated with the one-class label.
For these two models, decreasing the attack strength for generating the NRF dataset to $\epsilon=4/255$ produces less useful NRFs, but still achieve non-trivial performance.
However, EfficientNetB0 models achieve flipped results, obtaining random performance for $\epsilon=0.5, 0.25$ and non-trivial performance for $\epsilon=4/255$.

There are a few conclusions that we draw.
First, NRF usefulness seems to be \textit{independent of model size}, because the smallest and biggest model have useful NRFs while the medium-sized model does not.
Furthermore, as the attack strength decreases, the usefulness of NRFs generated decrease for useful NRFs and increase for useless NRFs.
Second, usefulness of NRFs learnt by models seems to be \textit{independent of the models' adversarial robustness}
-- ResNet18 is consistently the most robust to adversarial attacks across all subclasses and learns useful NRFs, while MobileNetV3 models learn useful NRFs and EfficientNetB0 models learn NRFs which are not useful.
Third,
the results of EfficientNetB0 highlight that the predictive power of one-class classifiers could come from learning some features that are not useful for the task, and their adversarial vulnerability arises from these learnt features being non-robust.

We also evaluate NRF models on the original test dataset and report the overall average precision in Fig. \ref{fig:NRF_Model_on_original_data} and subclass results in Tab. \ref{tab:nrf_models} in Appendix \ref{appendix:detailed results}.
Across all models and (sub)classes, we observe random performance of NRF models for NRF data attack strength $\epsilon=0.5$, even for the MobileNetV3 and ResNet18 NRF models trained with useful NRFs.
Across each model, the overall performance increases as the attack strength to generate the NRF data decreases, but subclass performance varies and the performance across (sub)classes is generally far lower than the models trained on the original dataset.
A difference in performance on the NRF and original test dataset suggests that NRF models are learning features apart from the NRFs from the PGD attack used to create the NRF datasets.
Moreover,
the drop from non-trivial performance on NRF data to almost random for the relevant NRF models hints that NRF models are learning features that are negatively correlated with the original one-class task.
The undesired consequence of learning features that are not useful (and potentially harmful) in standard models is echoed in NRF models too.
\looseness=-1

\begin{figure}
\footnotesize
\centering
\begin{subfigure}[t]{0.49\linewidth}
\centering
    \includegraphics[width=\linewidth]{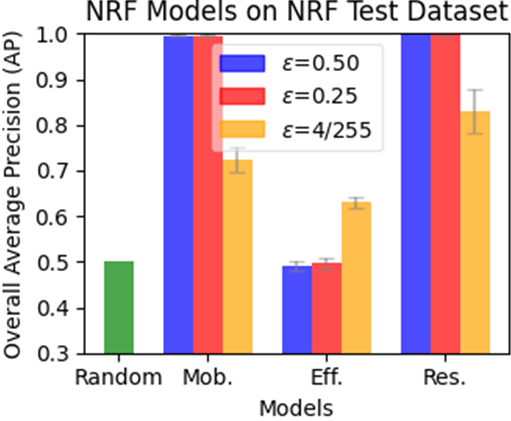}
    \caption{Tested on the NRF dataset.\label{fig:NRF_Model_on_NRF_data}}
\end{subfigure}
\begin{subfigure}[t]{0.49\linewidth}
\centering
    \includegraphics[width=\linewidth]{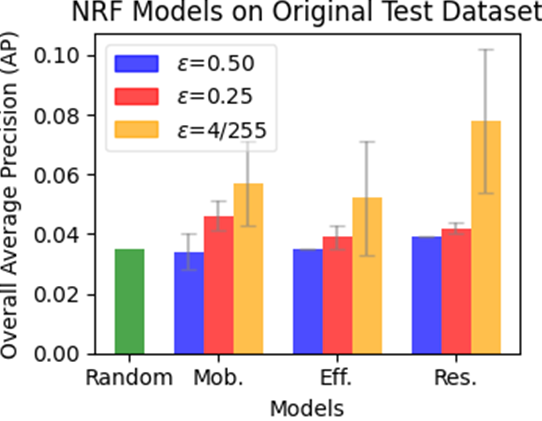}
    \caption{Tested on the original dataset.\label{fig:NRF_Model_on_original_data}}
\end{subfigure}
    \caption{Overall performance of NRF models trained on the non-robust feature (NRF) dataset, tested on the NRF and original dataset respectively.
    NRF datasets generated with $\ell_\infty$ PGD with varying strengths $\epsilon=0.5,0.25, 4/255$.}
    \label{fig:NRF_Model}
\end{figure}

\section{Conclusion}
In conclusion, we observe that one-class classifiers can still be vulnerable to adversarial attacks, but it may not be due to learning useful non-robust features.
We showed that one-class classifiers generally learn useful features for both seen and unseen data, but features learnt by smaller models (MobileNetV3-small) were less useful than bigger models as adversarial drift increases.
Moreover, successful attacks need to be stronger than previous multi-class classification settings, especially for bigger models (ResNet18).
Nevertheless, the presence of a successful attack suggests that these lightweight one-class classifiers still use non-robust features.
Unlike in \citet{adv_ex_features_not_bugs}, we show that the non-robust features learnt are sometimes not useful for the one-class task the model was trained for.
Furthermore, we show that model size and adversarial robustness are not good predictors on their usefulness.
These results show that model training can produce features that are not useful, not robust or both during training.
An important follow-up work would be to investigate the cause of models learning these unwanted features which are not useful nor robust, and how to prevent this.

{
    \small
    \bibliographystyle{ieeenat_fullname}
    \bibliography{bib}
}

\newpage
\appendix
\section{Training Details}

\subsection{Reproducibility}

In the following sections, we summarize some results to encourage the reproducibility of our results.
We will release the code on GitHub once anonymity issues are not a concern anymore.

\subsection{One-Class Classifier}
We design our one-class classifier to have a classification head on top of a frozen pre-trained encoder.
The classification head is trainable and designed for supervised anomaly detection, where the training data has the class of interest as well as examples from a proper subset of negative subclasses encountered during deployment.
We use the idea from \citet{lau2024OSR}, which is a neural network designed to enclose decision regions in the spirit of anomaly detection.
We detail our training details that follow their work.
We train a neural network with 2 leaky rectified linear unit (ReLU) layers and 1 linear layer with a Gaussian bump activation, with the last layer of a Gaussian radial basis function (RBF) having a fixed parameter at the all ones vector.
All linear layers maintain the same dimensionality as the previous layer, which amounts to $r\times r$ weight matrices where $r$ is the dimension of the penultimate layer of the encoder.

In this model, the features of the model would correspond to the output of the the bump activation, which represents the distance between the latent representation and a learnt hyperplane..
More precisely, since the output of the bump activation is between 0 and 1, we can translate it by any threshold to recover the formulation in Eq. (\ref{eqn:usefulness}).
Alternatively, one can replace the formulation in Eq. (\ref{eqn:usefulness}) to generalize the notion of correlation to arbitrary functions as well.

We fine-tune using logistic loss (binary cross entropy) with the Adam optimizer at a learning rate of 0.001 for 15 epochs with early stopping patience of 7.
From our experiments, 15 epochs allows good training and validation performance.

\subsection{Dataset}

Tab. \ref{tab:train test split} is the train-test split that we used for our experiments.
For CIFAR-10 classes, we randomly chose 300 images from each class for evaluation.
In our training set-up, we designed the one-class classification supervision to be balanced to allow training to be more effective, as compared to having few negatives.

\begin{table*}[]
    \centering
    \caption{Train-test split of data.}
    \begin{tabular}{l|l|c|c}
    \toprule
        Dataset & Class & Number of Training Samples & Number of Testing Samples  \\
        \midrule
        Drone vs. Bird & Drones (Positive) & 328& 100 \\
        \midrule
        Drone vs. Bird & Birds & 110 & 290 \\
        ImageNet-1K & Bald Eagle & 109 & 182\\
        ImageNet-1K & Airliner & 109 & 291 \\
        ImageNet-1K & Vulture & 0 & 295 \\
        ImageNet-1K & Airship & 0 & 323 \\
        ImageNet-1K & Crane2 & 0 & 304 \\
        ImageNet-1K & Jeep & 0 & 308 \\
        CIFAR-10 & Airplane & 0 & 300 \\
        CIFAR-10 & Bird & 0 & 300 \\
        CIFAR-10 & Truck & 0 & 300 \\
        \bottomrule
    \end{tabular}
    \label{tab:train test split}
\end{table*}

Since the encoder has been pre-trained on ImageNet-1K, the encoder has trained with an upstream task on the ImageNet-1K images we use during testing.
One potential limitation is the concern of test leakage for ImageNet-1K evaluations.
We opted to use ImageNet-1K in our evaluations because it is the most similar to the Drone vs. Bird dataset in terms of resolution, and has readily available labels.
Nevertheless, we believe that the effects of test leakage are mitigated for the following reason.
We ensured that we tested on other datasets: the Drone vs. Bird and CIFAR-10 datasets.
Our results from the Drone vs. Bird dataset show results on fine-tuned representations, while ImageNet-1K results show results on pre-trained and fine-tuned representations.
On the other hand, CIFAR-10 shows results on zero-shot representations.
Showing that one-class classifiers can achieve non-trivial performance across all negative subclasses (ImageNet-1K or not) suggests that our results on ImageNet-1K is not an overly optimistic estimate.
In fact, crane2 and jeep classes are seen during pre-training but still have worse performance than CIFAR-10 classes, which merely emphasizes the impact of semantic adversarial drift on the usefulness of features learnt.
As suggested by \citet{adv_ex_not_real_features}, this could be due to the fact that NRFs transfer poorly to other tasks.

\subsection{Adversarial Attacks}

To adversarially attack a model for robust performance evaluation and NRF dataset generation, we use 100 epochs with the same training hyperparameters for model training.
We use a step size of $\alpha=0.1$, as per \citet{adv_ex_features_not_bugs}.
\section{Detailed Results}
\label{appendix:detailed results}

\begin{figure*}[h]
    \centering
    \includegraphics[width=\textwidth]{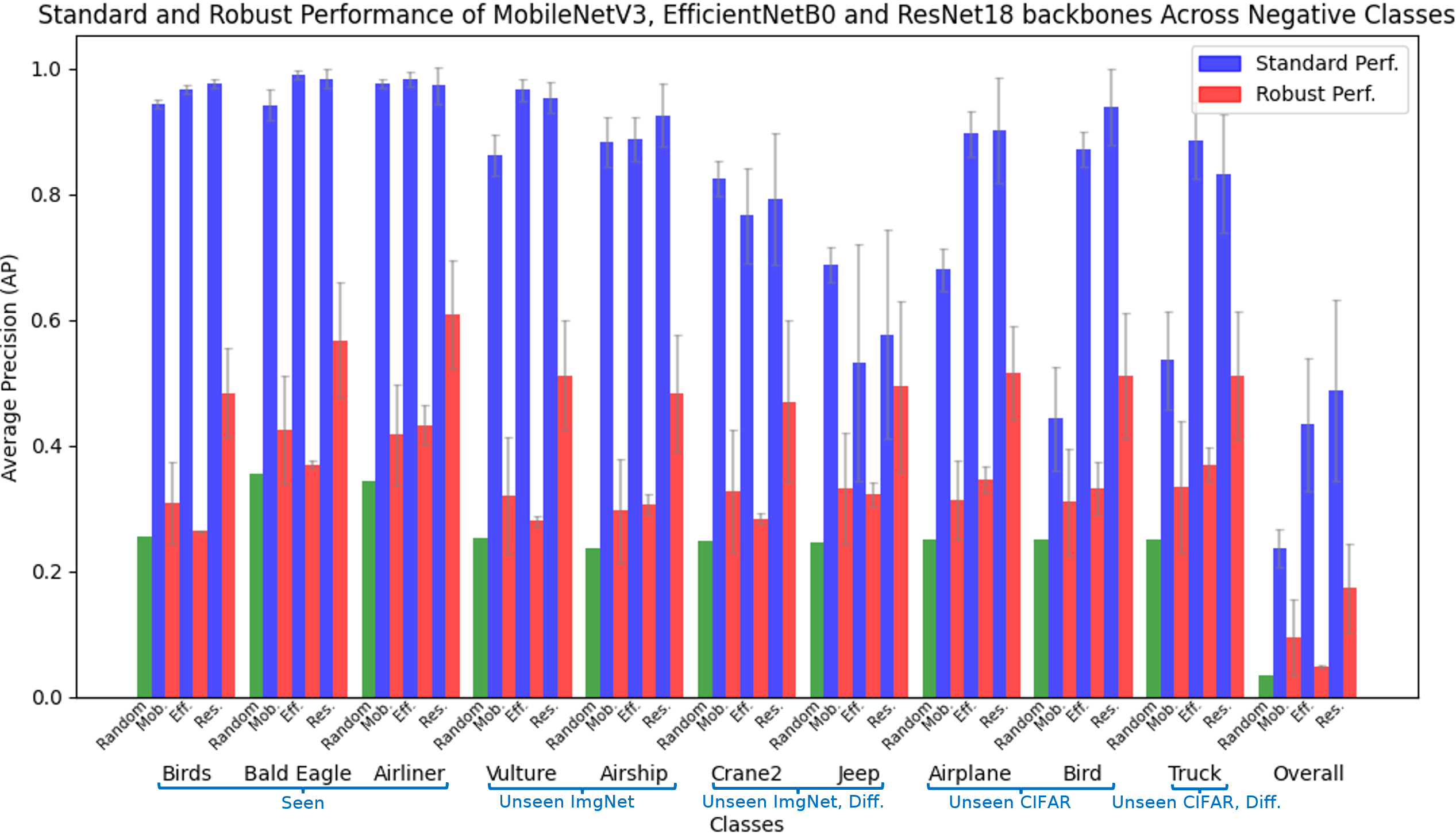}
    \caption{Standard and robust performance of different backbones across a subset of negative subclasses and the overall negative class.
    Each group of bars represents the average precision of the negative (sub)class against the positive drone class for different classifiers.
    The first green bar is the random baseline (calculated in expectation), followed by pairs of bars which are the standard and robust performance of each model of the following order: MobileNetV3 small (Mob.), EfficientNetB0 (Eff.) and ResNet18 (Res.).
    Classes are grouped by approximately increasing adversarial drift from left to right, with semantic differences (diff.) and imagery differences (ImageNet-1K/CIFAR-10).}
    \label{fig:full_standard_robust_perf}
\end{figure*}

In Fig. \ref{fig:full_standard_robust_perf}, we show the full version of the results across all subclasses shown in Fig. \ref{fig:standard_perf_pgd_linf_025}.
Tab. \ref{tab:drone_AUPR} has the corresponding numerical results, as well as performance of NRF models on the original test data.
Tab. \ref{tab:nrf_models} contains the performance of NRF models on the NRF test dataset.

\begin{table*}[t]
\addtolength{\tabcolsep}{-4pt}
\scriptsize
    \caption{Average precision of the positive class (drone) against seen classes (birds, bald eagle, airliner) and unseen classes (vulture, airship, jeep and crane2).
    Overall average precision is also reported.
    Random classifier is reported in expectation.
    We report the standard performance, robust performance and performance of NRF models trained with NRF datasets generated with $\ell_\infty$ PGD under varying strengths $\epsilon=0.5,0.25, 4/255$.
        \looseness=-1
        }
        \centering
    \begin{tabular}{cl|ccc|cccc|ccc|c}
        \toprule
        & & \multicolumn{3}{c|}{Seen} & \multicolumn{4}{c|}{Unseen ImageNet} & \multicolumn{3}{c|}{Unseen CIFAR-10} & \\
\multicolumn{2}{c|}{Model$\backslash$Class}   & Birds  & Bald Eagle  & Airliner  & Vulture & Airship & Crane2  & Jeep & Airplane & Bird & Truck & Overall     \\
        \midrule
 &Random  &  0.256 & 0.355 & 0.344 & 0.253 & 0.236 & 0.248 & 0.245 & 0.250 & 0.250 & 0.250 &  0.035 \\
 \midrule
\parbox[t]{2mm}{\multirow{5}{*}{\rotatebox[origin=c]{90}{MobileNetV3}}} & Standard & 0.943$\pm$0.007 & 0.942$\pm$0.024 & 0.977$\pm$0.007 & 0.862$\pm$0.032 & 0.883$\pm$0.040 & 0.825$\pm$0.027 & 0.688$\pm$0.027 & 0.680$\pm$0.033 & 0.443$\pm$0.082 & 0.536$\pm$0.078 & 0.237$\pm$0.030\\
&Robust & 0.309$\pm$0.065 & 0.424$\pm$0.086 & 0.417$\pm$0.081 & 0.321$\pm$0.093 & 0.296$\pm$0.083 & 0.327$\pm$0.098 & 0.332$\pm$0.088 & 0.314$\pm$0.063 & 0.310$\pm$0.085 & 0.335$\pm$0.105 & 0.095$\pm$0.060\\
&NRF ($\epsilon$=0.5) & 0.228$\pm$0.019 & 0.385$\pm$0.104 & 0.462$\pm$0.021 & 0.271$\pm$0.062 & 0.250$\pm$0.029 & 0.340$\pm$0.065 & 0.518$\pm$0.115 & 0.209$\pm$0.041 & 0.170$\pm$0.029 & 0.289$\pm$0.121 & 0.034$\pm$0.006 \\
&NRF ($\epsilon$=0.25) &  0.276$\pm$0.049 & 0.369$\pm$0.026 & 0.350$\pm$0.027 & 0.256$\pm$0.010 & 0.223$\pm$0.030 & 0.282$\pm$0.038 & 0.373$\pm$0.157 & 0.489$\pm$0.129 & 0.505$\pm$0.116 & 0.681$\pm$0.151 & 0.046$\pm$0.005\\
&NRF ($\epsilon$=4/255) &  0.326$\pm$0.031 & 0.430$\pm$0.089 & 0.434$\pm$0.095 & 0.312$\pm$0.094 & 0.308$\pm$0.074 & 0.315$\pm$0.060 & 0.286$\pm$0.030 & 0.517$\pm$0.173 & 0.574$\pm$0.272 & 0.575$\pm$0.285 & 0.057$\pm$0.014\\
 \midrule
\parbox[t]{2mm}{\multirow{5}{*}{\rotatebox[origin=c]{90}{EfficientNetB0}}} & Standard & 0.966$\pm$0.007 & 0.990$\pm$0.006 & 0.983$\pm$0.011 & 0.966$\pm$0.018 & 0.888$\pm$0.034 & 0.766$\pm$0.076 & 0.532$\pm$0.188 & 0.896$\pm$0.036 & 0.872$\pm$0.028 & 0.885$\pm$0.061 & 0.434$\pm$0.106\\
 & Robust & 0.264$\pm$0.001 & 0.370$\pm$0.006 & 0.433$\pm$0.031 & 0.280$\pm$0.008 & 0.306$\pm$0.017 & 0.284$\pm$0.008 & 0.323$\pm$0.018 & 0.346$\pm$0.022 & 0.331$\pm$0.042 & 0.370$\pm$0.028 & 0.048$\pm$0.003 \\
&NRF ($\epsilon$=0.5) & 0.256$\pm$0.000 & 0.355$\pm$0.000 & 0.344$\pm$0.000 & 0.253$\pm$0.000 & 0.236$\pm$0.000 & 0.248$\pm$0.000 & 0.245$\pm$0.000 & 0.250$\pm$0.000 & 0.250$\pm$0.000 & 0.250$\pm$0.000 & 0.035$\pm$0.000 \\
&NRF ($\epsilon$=0.25) &  0.288$\pm$0.003 & 0.353$\pm$0.007 & 0.376$\pm$0.012 & 0.249$\pm$0.007 & 0.270$\pm$0.012 & 0.267$\pm$0.008 & 0.243$\pm$0.007 & 0.288$\pm$0.013 & 0.288$\pm$0.013 & 0.288$\pm$0.013 & 0.039$\pm$0.004\\
&NRF ($\epsilon$=4/255) &  0.268$\pm$0.065 & 0.526$\pm$0.088 & 0.522$\pm$0.033 & 0.385$\pm$0.067 & 0.313$\pm$0.063 & 0.343$\pm$0.052 & 0.242$\pm$0.095 & 0.411$\pm$0.153 & 0.447$\pm$0.174 & 0.462$\pm$0.255 & 0.052$\pm$0.019\\
\midrule
\parbox[t]{2mm}{\multirow{5}{*}{\rotatebox[origin=c]{90}{ResNet18}}} & Standard & 0.975$\pm$0.007 & 0.984$\pm$0.016 & 0.973$\pm$0.029 & 0.954$\pm$0.025 & 0.926$\pm$0.050 & 0.793$\pm$0.105 & 0.577$\pm$0.167 & 0.902$\pm$0.084 & 0.939$\pm$0.060 & 0.833$\pm$0.095 & 0.488$\pm$0.144\\
&Robust & 0.484$\pm$0.071 & 0.568$\pm$0.093 & 0.609$\pm$0.086 & 0.512$\pm$0.088 & 0.482$\pm$0.093 & 0.470$\pm$0.129 & 0.494$\pm$0.136 & 0.516$\pm$0.074 & 0.511$\pm$0.101 & 0.511$\pm$0.103 & 0.173$\pm$0.071\\
&NRF ($\epsilon$=0.5) & 0.224$\pm$0.028 & 0.333$\pm$0.041 & 0.430$\pm$0.032 & 0.286$\pm$0.034 & 0.246$\pm$0.006 & 0.249$\pm$0.051 & 0.362$\pm$0.162 & 0.339$\pm$0.079 & 0.364$\pm$0.084 & 0.448$\pm$0.120 & 0.039$\pm$0.000 \\
&NRF ($\epsilon$=0.25) &  0.475$\pm$0.022 & 0.333$\pm$0.038 & 0.380$\pm$0.012 & 0.278$\pm$0.028 & 0.289$\pm$0.011 & 0.200$\pm$0.011 & 0.172$\pm$0.011 & 0.672$\pm$0.043 & 0.678$\pm$0.065 & 0.660$\pm$0.083 & 0.042$\pm$0.002\\
&NRF ($\epsilon$=4/255) &  0.496$\pm$0.106 & 0.529$\pm$0.062 & 0.497$\pm$0.173 & 0.446$\pm$0.027 & 0.533$\pm$0.086 & 0.320$\pm$0.078 & 0.270$\pm$0.073 & 0.538$\pm$0.275 & 0.554$\pm$0.294 & 0.508$\pm$0.294 & 0.078$\pm$0.024\\
        \bottomrule
    \end{tabular}
    \label{tab:drone_AUPR}
\end{table*}

\begin{table*}[t]
\addtolength{\tabcolsep}{-4pt}
\scriptsize
    \caption{Performance of models trained with non-robust features (NRFs) on the NRF test data.
    Models tested have MobileNetV3-small, EfficientNetB0 and ResNet18 backbones.
    NRFs are generated with $\ell_\infty$ PGD with varying strengths of $\epsilon=0.5, 0.25, 4/255$, according to the respective model backbone during standard training.
    Average precision of the positive class (drone) against seen classes (birds, bald eagle, airliner) and unseen classes (vulture, airship, jeep and crane2).
    Overall average precision is also reported.
    Random classifier is reported in expectation.
    MobileNetV3 and ResNet18 models achieve close to perfect performance on $\epsilon=0.5,0.25$, suggesting that non-robust features learnt during standard training are useful.
    However, EfficientNetB0 achieves random performance for these attack strengths, suggesting that NRFs learnt during standard training are not useful.
        \looseness=-1
        }
        \centering
    \begin{tabular}{cl|ccc|cccc|ccc|c}
        \toprule
         && \multicolumn{3}{c|}{Seen} & \multicolumn{4}{c|}{Unseen ImageNet} & \multicolumn{3}{c|}{Unseen CIFAR-10} & \\
\multicolumn{2}{c|}{Model$\backslash$Class}    & Birds  & Bald Eagle  & Airliner  & Vulture & Airship & Crane2  & Jeep & Airplane & Bird & Truck & Overall     \\
        \midrule
 &Random  &  0.500 & 0.500 & 0.500 & 0.500 & 0.500 & 0.500 & 0.500 &  0.500 & 0.500 & 0.500 &  0.500 \\
 \midrule
\parbox[t]{2mm}{\multirow{3}{*}{\rotatebox[origin=c]{90}{Mob.}}} &$\epsilon$=0.5 & 0.990$\pm$0.014 & 0.993$\pm$0.007 & 0.995$\pm$0.007 & 0.995$\pm$0.005 & 0.993$\pm$0.011 & 0.996$\pm$0.004 & 0.997$\pm$0.005 & 0.994$\pm$0.010 & 0.993$\pm$0.011 & 0.990$\pm$0.012 & 0.996$\pm$0.003 \\
&$\epsilon$=0.25 & 0.989$\pm$0.005 & 0.998$\pm$0.002 & 0.998$\pm$0.003 & 0.999$\pm$0.001 & 0.998$\pm$0.001 & 0.997$\pm$0.003 & 0.999$\pm$0.001 & 0.996$\pm$0.002 & 0.993$\pm$0.006 & 0.995$\pm$0.004 & 0.996$\pm$0.001 \\
&$\epsilon$=4/255 & 0.737$\pm$0.078 & 0.758$\pm$0.056 & 0.811$\pm$0.051 & 0.774$\pm$0.051 & 0.794$\pm$0.032 & 0.785$\pm$0.034 & 0.854$\pm$0.014 & 0.679$\pm$0.050 & 0.658$\pm$0.041 & 0.694$\pm$0.042 & 0.724$\pm$0.027 \\
 \midrule
\parbox[t]{2mm}{\multirow{3}{*}{\rotatebox[origin=c]{90}{Eff.}}}&$\epsilon$=0.5 & 0.508$\pm$0.028 & 0.450$\pm$0.039 & 0.494$\pm$0.032 & 0.467$\pm$0.025 & 0.469$\pm$0.033 & 0.511$\pm$0.007 & 0.511$\pm$0.047 & 0.506$\pm$0.021 & 0.472$\pm$0.036 & 0.490$\pm$0.008 & 0.491$\pm$0.011 \\
&$\epsilon$=0.25 & 0.500$\pm$0.018 & 0.454$\pm$0.049 & 0.484$\pm$0.048 & 0.449$\pm$0.011 & 0.476$\pm$0.018 & 0.499$\pm$0.022 & 0.513$\pm$0.063 & 0.506$\pm$0.012 & 0.486$\pm$0.009 & 0.498$\pm$0.023 & 0.496$\pm$0.013 \\
&$\epsilon$=4/255 & 0.637$\pm$0.023 & 0.591$\pm$0.016 & 0.647$\pm$0.010 & 0.647$\pm$0.036 & 0.669$\pm$0.059 & 0.704$\pm$0.037 & 0.680$\pm$0.094 & 0.613$\pm$0.074 & 0.592$\pm$0.061 & 0.597$\pm$0.057 & 0.631$\pm$0.012 \\
\midrule
\parbox[t]{2mm}{\multirow{3}{*}{\rotatebox[origin=c]{90}{Res.}}}& $\epsilon$=0.5 & 1.000$\pm$0.000 & 1.000$\pm$0.000 & 1.000$\pm$0.000 & 1.000$\pm$0.000 & 1.000$\pm$0.000 & 1.000$\pm$0.000 & 1.000$\pm$0.000 & 1.000$\pm$0.000 & 1.000$\pm$0.000 & 1.000$\pm$0.000 & 1.000$\pm$0.000 \\
&$\epsilon$=0.25 & 0.999$\pm$0.002 & 0.997$\pm$0.002 & 0.999$\pm$0.001 & 0.998$\pm$0.004 & 0.999$\pm$0.001 & 0.998$\pm$0.002 & 0.998$\pm$0.002 & 0.997$\pm$0.004 & 0.998$\pm$0.002 & 0.998$\pm$0.002 & 0.998$\pm$0.002 \\
&$\epsilon$=4/255 & 0.845$\pm$0.039 & 0.874$\pm$0.025 & 0.843$\pm$0.093 & 0.883$\pm$0.041 & 0.824$\pm$0.063 & 0.848$\pm$0.050 & 0.864$\pm$0.087 & 0.797$\pm$0.056 & 0.818$\pm$0.038 & 0.828$\pm$0.052 & 0.830$\pm$0.049 \\
        \bottomrule
    \end{tabular}
    \label{tab:nrf_models}
\end{table*}

\end{document}